\documentclass{svproc}
\usepackage{url}

\usepackage{amsmath}
\usepackage{graphicx}
\usepackage{todonotes}
\usepackage[para]{threeparttable}
\usepackage{comment}

\usepackage{xcolor}
\definecolor{verylightgray}{gray}{0.9}

\usepackage{listings}
\usepackage[nolist]{acronym}
\acrodef{llm}[LLM]{large language model}
\acrodef{rps}[RPS]{Robot Programming Suite}
\acrodef{ide}[IDE]{integrated development environment}
\acrodef{sme}[SME]{small- or medium-sized enterprise}
\acrodef{gpu}[GPU]{graphics processing unit}
\acrodef{rlhf}[RLHF]{Reinforcement Learning with Human Feedback}

\newcommand{\forceindent}{\leavevmode{\parindent=1.5em\indent}}

\begin{document}
\mainmatter              % start of a contribution
\title{Domain-Specific Fine-Tuning of Large Language Models for Interactive Robot Programming}
\titlerunning{Domain-Specific Fine-Tuning of LLMs}  % abbreviated title (for running head)
%                                     also used for the TOC unless
%                                     \toctitle is used
%
\author{Benjamin Alt\inst{1} \and Urs Keßner\inst{1,2} \and Aleksandar Taranovic\inst{2} \and Darko Katic\inst{1} \and Andreas Hermann\inst{1} \and Rainer Jäkel\inst{1} \and Gerhard Neumann\inst{2}}
\authorrunning{Benjamin Alt et al.} % abbreviated author list (for running head)
%
%%%% list of authors for the TOC (use if author list has to be modified)
\tocauthor{Benjamin Alt, Urs Keßner, Darko Katic, Andreas Hermann, Rainer Jäkel and Gerhard Neumann}
\institute{ArtiMinds Robotics, 76131 Karlsruhe, Germany,\\
\email{benjamin.alt@artiminds.com}
\and
Karlsruhe Institute of Technology,
Autonomous Learning Robots Lab,\\
76131 Karlsruhe, Germany}

\maketitle              % typeset the title of the contribution

\setcounter{footnote}{0} 

\begin{abstract}
Industrial robots are applied in a widening range of industries, but robot programming mostly remains a task limited to programming experts. We propose a natural language-based assistant for programming of advanced, industrial robotic applications and investigate strategies for domain-specific fine-tuning of foundation models with limited data and compute.

\end{abstract}
\section{Introduction}
Industrial robot programming is cumbersome and costly, especially for advanced applications requiring force sensing and vision. Skill-based approaches simplify programming complex robot tasks by sequencing learned or pre-programmed sub-skills such as grasping or peg-in-hole fitting. However, skill-based programming this still requires expert knowledge, e.g., physical dependencies between acceleration and contact forces, but also technical aspects, such as the applicability of skills in collision-prone or low-tolerance environments. With the introduction of large-scale pretrained foundation models such as LLaMA \cite{touvron_llama_2023}, there is an emerging body of work on fine-tuning \acp{llm} for use in various application domains \cite{chalkidis_legal-bert_2020,zheng_trafficsafetygpt_2023}.
We propose a language-based programming assistant, which offers interactive dialogues about skills, example use-cases and expected robot behavior. We present three alternative model families, trained by three different domain-specific fine-tuning approaches. The variants are evaluated by comparing their BERTScore \cite{zhang_bertscore_2020} performance as well as a user survey with industry experts. 

\begin{comment}
\section{Related Work}
% \subsection{Interactive Robot Programming Assistants}

 Recent works like PaLM-E (Driess et al., 2023) incorporate LLMs and vision-language models (VLMs) (Alayrac et al., 2022; Chen et al., 2023) into robotics.
 However, they mimic a state machine which parses commands into  primitives, which are executed on a low-level controller. To address shortcomings of VLMs, Brohan et al. (2023) introduce a method to directly train VLMs designed for open-vocabulary visual question answering and visual dialogue to output low-level robot actions. They adapted two pre-trained VLMs, PaLM-E (Driess et al., 2023) and PaLI-X (Chen et al., 2023) on robotic trajectories by tokenizing the actions into text tokens. Brohan et al. (2023) define this new model category as vision-language-action (VLA) models. They based the action encoding of the discretization of Brohan et al. (2023). However, this approach lacks clear, easy to understand assistance, which guides even novice programmers through the process of setting up complex robot programs.
%% What is the distinction to our apporach? I just copied this from Urs,
\end{comment}

\section{Methods}

As a testbed, we use the ArtiMinds \ac{rps}, an \ac{ide} for industrial robots. With it, robots are programmed using parameterizable robot skills (called ``templates'') such as ``Grasp'', ``Insert'', etc. The resulting robot program is a tree of parameterized templates, which is compiled into executable code. The assistant answers questions of three types:

\textbf{Providing high-level explanations of templates.} Given a question like ``What does a `Move to State' template do?'', it should answer something like ``It moves the end-effector on a collision-free trajectory to a goal specified in configuration space.'' The complexity of answers range from simple, as illustrated here, to highly complex, particularly for force-controlled skills whose behavior depends in part on interactions with the environment.

\textbf{Providing examples for the usage of templates.} To a question like ``When should I use a `Move to State' template?'', it should answer ``A `Move to State' template can be used to efficiently move the robot arm through environments in which collisions can occur.'' Examples can involve concrete application domains, such as painting, gluing or welding for a ``Path Loader'' template.

\textbf{Providing step-by-step explanations of expected robot behavior.} To a question like ``What motions will a robot make when executing a `Grasp' template?'', the assistant should answer ``It will execute a collision-free approach motion, open the gripper, move closer to the object, close the gripper, and depart with a collision-free motion.'' Step-by-step explanations provide useful information for very high-level templates such as ``Peg in Hole''.

We investigate how \acp{llm} can be fine-tuned under the compute and data constraints typical for industrial \acp{sme}. This implies limitations to a single server-grade \ac{gpu}, e.g., Nvidia A100, with 80GB VRAM. Traditional fine-tuning requires up to 780GB given 65B parameters \cite{dettmers_qlora_2023}. To reduce model size, we leverage QLoRA adaptation with 4-bit NormalFloat quantization and double quantization \cite{dettmers_qlora_2023}. We investigate three data sparse training regimes: Fine-tuning an instruction-following model on domain-specific data, which has been pretrained in different domains; fine-tuning a streaming model for instruction-following on domain-specific data; and fine-tuning a streaming model on a domain-specific streaming data, succeeded by domain-specific instruction-following training. Comparing these training regimes determines whether the use of general-purpose pre-trained models provides useful priors to simplify fine-tuning. Moreover, we investigate the impact of prefix fine-tuning \cite{li_prefix-tuning_2021} on model quality.

\subsection{Datasets}

\forceindent \textbf{Streaming dataset:}
%% \begin{figure}
%%  \centering
%%    \begin{lstlisting}[breakindent=0pt]
%% "Editing paths: ArtiMinds RPS allows us to compute or record complex paths as robot trajectories. You can manipulate those trajectories with different path-blending parameters and change acceleration and velocity settings according to the capabilities of the configured robot manufacturer."
%%  \end{lstlisting}
%%  \caption{Extract of $\mathcal{D}_{stream}$ (streaming dataset).  }
  %%\todo[inline]{RFC: Which figures should be cut?}
%%  \label{fig:streaming-dataset}
%% \end{figure}
One of the investigated training regimes involves the fine-tuning of a general-purpose streaming model such as LLaMA on domain-specific data. We create a dataset $\mathcal{D}_{stream}$ by parsing the ArtiMinds \ac{rps} user manual into a plain-text representation. Non-informational elements like the cover page, table of contents, index pages, and copyright paragraphs were removed to focus on the technical content.
\textbf{Instruction-following datasets:}
We constructed a domain-specific dataset to evaluate whether a model could be trained to answer questions using only natural language instructions. The dataset contains questions about the usage of ArtiMinds \ac{rps} templates, asking for template descriptions, use cases, and step-by-step descriptions extracted from the \ac{rps} documentation. Zhang and Soh \cite{zhang_large_2023} found that models may be overly sensitive to minor differences in input prompts, resulting in unintended variability in the generated responses. To address this, we generated 10 question variants for each information topic using ChatGPT 3.5 \cite{openai_chatgpt_2023}.  The resulting dataset $\mathcal{D}_{instr}$ contains a total of 250 instruction-label pairings. % (see Fig. \ref{fig:instruction-dataset}).

\textbf{Prefix fine-tuning} is a technique in which a specific prefix is added to each prompt to guide the model into the correct domain \cite{li_prefix-tuning_2021}. To determine how this affects domain-specific fine-tuning, we created an additional version $\mathcal{D}_{instr}^{prefix}$ of the instruction-following dataset, with a custom, domain-specific prefixes, while $\mathcal{D}_{instr}$ applies the default Alpaca prefix \cite{taori_alpaca_2023}. % An example instruction from $\mathcal{D}_{instr}^{prefix}$ is shown in Fig. \ref{fig:instruction-dataset}.

\subsection{Model Variants}

\forceindent \textbf{Alpaca Finetuned on Instruction Task:}
\label{sec:finetuned-alpaca}
We finetune three pretrained Alpaca models on our domain-specific instruction-following dataset $\mathcal{D}_{instr}$. The 7B and 13B models were fully fine-tuned (see Taori et al.), while the 30B model was finetuned using QLoRA, and the resulting LoRA adapters merged back into the original foundation model. Double 4-bit NormalFloat quantization \cite{dettmers_qlora_2023} is applied. We perform QLoRA with rank $r=8$, $\alpha=16$ and $\text{dropout}=0.05$ for all 4 self-attention matrices.

\textbf{LLaMA Finetuned on Instruction Task:}
\label{sec:instruction-llama}
Instead of fine-tuning a pretrained instruction-following model, domain adaptation could also be achieved by fine-tuning a streaming model on an instruction-following dataset in the new domain. We use LLaMA models (7B, 13B and 30B) and quantize them to double-quantized 4-bit NormalFloat. We train a LoRA adapter on the instruction-following task $\mathcal{D}_{instr}$, using the same hyperparameters as in Section \ref{sec:finetuned-alpaca}.

\textbf{Merged LLaMA (Streaming \& Instruction Task):} \label{sec:merged-llama} A third variant of domain adaptation is to perform domain-specific fine-tuning in two stages: First on a streaming task, and then on an instruction-following task, both in the new domain.
Using the same pretrained and quantized LLaMA models as in Section \ref{sec:instruction-llama}, we first train a LoRA adapter on the streaming dataset $\mathcal{D}_{stream}$. Then, we train an additional LoRA adapter on $\mathcal{D}_{instr}$.

\textbf{Prefix Fine-tuning} To assess to what extent prefix fine-tuning facilitates domain adaptation, we train additional variants of all models mentioned above on $\mathcal{D}_{stream}^{prefix}$ and $\mathcal{D}_{instr}^{prefix}$ respectively.

\section{Experiments}

\subsection{BERTScore Evaluation}
To assess the semantic correctness of the trained models' responses, we compute BERTScores \cite{zhang_bertscore_2020} with an evaluation dataset. Nine instructions and reference responses to questions about RPS templates were manually written by domain experts. BERTScore recall, precision and F1 scores were calculated between each response and its reference.

The results are reported in Table \ref{tab:bertscore}. Alpaca models without prefix performed best with an $F_{BERT}$ of 0.8. We observe that directly fine-tuning a pretrained instruction-following model yields the best results, and that model size seems uncorrelated to BERTScore performance.

\begin{table}[t]
\caption{BERTScores and survey results for the 10 best-performing models, sorted by $F_{BERT}$. Factual correctness ($C$) and domain adherence ($D$) are binary features, perceived helpfulness ($H$) is ranked on a five-point Likert scale (1 - low, 5 - high).}
\label{tab:bertscore}
\setlength{\tabcolsep}{4pt}
\begin{center}
\begin{threeparttable}
\begin{tabular}{l|ccc|cccc}
\hline\rule{0pt}{12pt}%
& $R_{BERT}$ &   $P_{BERT}$ &   $F_{BERT}$ &       $C$ &      $D$ &     $H$ &   $N$\tnote{$\dagger$} \\[2pt]
\hline\rule{0pt}{12pt}%
Alpaca 30B             &       0.8156 &       0.7945 &       0.8042 & 0.3830 & 1.0000 & 3.0851 &     5 \\
 Alpaca 7B              &       0.8044 &       0.7979 &       0.8007 & 0.2041 & 0.8776 & 1.9796 &     5 \\
 Alpaca 13B w. p.\tnote{*}       &       0.7868 &       0.7929 &       0.7894 & 0.4000 & 0.8600 & 2.5000 &     5 \\
 Alpaca 7B w. p.        &       0.7873 &       0.7780 &       0.7818 & 0.2881 & 0.5085 & 2.4407 &     6 \\
 LLaMA 13B w. p.        &       0.8238 &       0.7421 &       0.7798 & 0.1800 & 0.7800 & 2.4600 &     5 \\
 Alpaca 30B w. p.       &       0.7722 &       0.7189 &       0.7438 & 0.0506 & 0.2785 & 1.3038 &     8 \\
 Merged LLaMA 7B w. p.  &       0.8017 &       0.6891 &       0.7386 & 0.2317 & 0.9146 & 2.2317 &     9 \\
 Merged LLaMA 30B       &       0.7952 &       0.6780 &       0.7311 & 0.1224 & 0.8776 & 2.0000 &     5 \\
 LLaMA 13B              &       0.7910 &       0.6747 &       0.7266 & 0.2143 & 0.6429 & 2.1714 &     7 \\
 LLaMA 30B              &       0.7631 &       0.6784 &       0.7174 & 0.1277 & 0.7872 & 1.8298 &     5 \\
[2pt]
\hline
\end{tabular}
\begin{tablenotes}
    \footnotesize
    \item[*] with prefix
    \item[$\dagger$] number of survey responses
\end{tablenotes}
\end{threeparttable}
\end{center}
\end{table}

\subsection{User Survey}

To obtain feedback about real-life performance, we conduct a survey with domain experts. The evaluation set contains 40 instructions, some of which outside the question types seen in training (e.g., ``What is the difference between a 'Move to Point' template and a 'Move to State' template?''). A total of 33 engineers participated in the survey, each completing at least one questionnaire containing 10 randomly assigned prompts and model responses. Each prompt-response pair was evaluated on factual correctness (``The answer is precise and factually correct. (yes/no)''), domain adherence (``The answer remains in the domain of ArtiMinds RPS. (yes/no)'') and perceived helpfulness (``Rate the helpfulness of the response on a scale from 1 to 5.'').
%65B models were excluded from the user study, as their training had not yet finished.
The results are shown in Table \ref{tab:bertscore}. Echoing the BERTScore results, Alpaca models performed best with respect to correctness, domain adherence and helpfulness.
%The high domain adherence ratings of Alpaca-based models in particular support the viability of domain-specific fine-tuning of an instruction-following model over the investigated alternative approaches.
It must be noted, however, that all trained models have noticeable shortcomings with respect to correctness and overall helpfulness. The LLaMA-based models in particular struggled with sentence structure, often repeating words and predicting unknown tokens. In contrast, Alpaca-based models generated responses with satisfactory format. Prefix fine-tuning did not have a discernible impact on model performance.

\section{Conclusions}

We propose a natural-language-based programming assistant which offers human-like, interactive dialog for assistance with complex industrial robot programming. The assistant offers explanations of skills, example use-cases and expected robot behavior. We train three families of \acp{llm} realizing three different approaches for domain-specific fine-tuning and evaluate them with respect to a quantitative semantic similarity metric as well as a user survey. The results indicate that domain-specific fine-tuning of instruction-following models achieves robust domain transfer. However, the results are not (yet) sufficient for practical use. The models often exhibit strong hallucinations (including out of domain responses) and repeat words or phrases or even switch from English to German. This suggests that fine-tuning alone is insufficient to imprint the new knowledge into the model with limited data. Prompting strategies, which eschew fine-tuning in favor of carefully designed prompts, are promising alternatives, and have become technically feasible with the most recent generation of foundation models. Likewise, larger-scale user surveys are required to draw more robust and nuanced conclusions about real-world model performance.

\section*{Acknowledgment}
This  work  was  supported  by  the  German  Federal  Ministry  of  Education and Research (grant 01DR19001B).

\bibliography{bibliography_alt}
\bibliographystyle{bibtex/splncs03}

\end{document}